# Robust language-based mental health assessments in time and space through social media

Siddharth Mangalik[a,1], Johannes C. Eichstaedt[b,c,1], Salvatore Giorgi[e], Jihu Mun[a], Farhan Ahmed[a], Gilvir Gill[a], Adithya V. Ganesan[a], Shashanka Subrahmanya[c], Nikita Soni[a], Sean A. P. Clouston[d], and H. Andrew Schwartz[a,1]

[a]Department of Computer Science, Stony Brook University, Stony Brook, NY, USA; [b]Department of Psychology, Stanford University, Stanford, CA, USA; [c]Institute for Human-Centered A.I., Stanford University, CA, USA; [d]Department of Family, Population, and Preventive Medicine, Renaissance School of Medicine, Stony Brook University, Stony Brook, NY, USA; [e]Department of Computer and Information Science, University of Pennsylvania

This manuscript was compiled on February 25, 2023

**Compared to physical health, population mental health measurement in the U.S. is very coarse-grained. Currently, in the largest population surveys, such as those carried out by the Centers for Disease Control or Gallup, mental health is only broadly captured through "mentally unhealthy days" or "sadness", and limited to relatively infrequent state or metropolitan estimates. Through the large scale analysis of social media data, robust estimation of population mental health is feasible at much higher resolutions, up to weekly estimates for counties. In the present work, we validate a pipeline that uses a sample of 1.2 billion Tweets from 2 million geo-located users to estimate mental health changes for the two leading mental health conditions, depression and anxiety. We find moderate to large associations between the language-based mental health assessments and survey scores from Gallup for multiple levels of granularity, down to the county-week (fixed effects $\beta = .25$ to $1.58$; $p < .001$). Language-based assessment allows for the cost-effective and scalable monitoring of population mental health at weekly time scales. Such spatially fine-grained time series are well suited to monitor effects of societal events and policies as well as enable quasi-experimental study designs in population health and other disciplines. Beyond mental health in the U.S., this method generalizes to a broad set of psychological outcomes and allows for community measurement in under-resourced settings where no traditional survey measures – but social media data – are available.**

Pre-Print | Depression | Anxiety | Social Media Analysis | Spatiotemporal

Mental health is a large public health concern, causing large economic impact and loss of quality of life. Recent estimates suggest that depression affects 19.4 million Americans (7.8% of the population, 2020 est.) each year (1), while generalized anxiety disorder affects approximately 6% of the US population (19.8 million people, 2010 est.) (2). Globally, mental health conditions are the fifth-most common cause of reduced quality of life (3). Critically, poor mental health is thought to play a central role driving recent increases in prevalence and severity of "deaths of despair" (4, 5) in part due to the influence of poorer mental health on suicide attempts and suicide mortality obesity (6), and opioid-related overdoses (7).

Public health researchers and policy makers seek to understand and actively respond to emerging and changing conditions (8, 9). Yet, current standards for monitoring mental health outcomes rely on subjective information from surveys that have limited temporal or regional resolution. For example, annual changes in depression are measured only by annual Gallup polling (10) and a handful of national surveys (11) while anxiety is not regularly assessed in any of these surveys (12). Nevertheless, improving geospatial resolution can provide researchers with tools to more reliably assess the distribution (13) and determinants of disease (14). Similarly, a wealth of small studies using ecological momentary assessment suggest that observations made on shorter timescales routinely identifies symptoms and correlates that are otherwise inaccessible to researchers (15, 16).

Applying validated measures of depression and anxiety, assessed objectively at regular time-intervals at the county-level could transform research in population mental health, allowing researchers for the first time to locate clusters and reasons for changes to poorer mental health (17). Since originally proposed, language-based assessments have developed to become a flexible source of objective information about individuals' emotions and behaviors (18), often with greater accuracy and predictive power than existing survey-based measures (19). Further, recent work has found significant increases in convergent validity via post-stratification techniques (20) to address known selection biases (21, 22).

Here, we bring integrate a series of recent advances into a single pipeline, *language-based mental health assessment* (*LBMHA*: Figure 1), to produce anxiety and depression estimates over regions and time. We first establish the reliability of this approach, contrasted with standard survey approaches, while varying the time and space units (from annual, national down to daily, townships) as well as minimum thresholds for time-space specific observations. We then evaluate the convergent and external validity of the measurements as compared to the most extensively collected surveys available, both cross-sectionally and longitudinally. To facilitate open scientific inquiry we provide both the LBMHA measures as well as an open-source toolkit for running the pipeline and deriving mental health estimates.

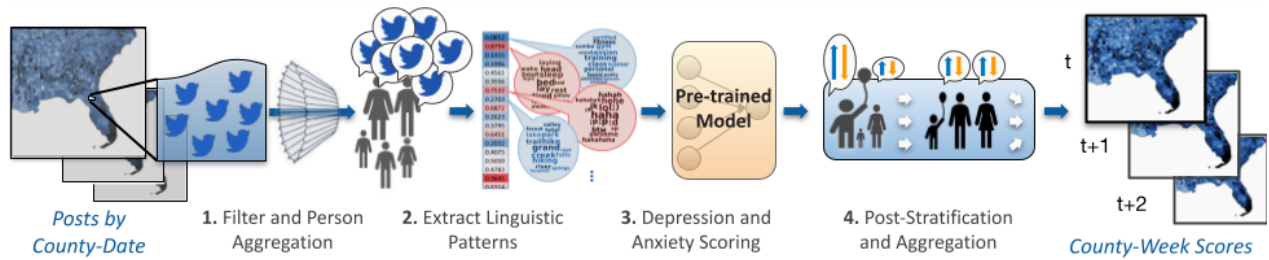

**Fig. 1.** A brief visual overview of how language is captured and tagged by county and week on social media platforms, an also explains how these data are then used to generate weighted depression and anxiety scores. County mapped messages are filtered to represent self-written language, the language extracted from these messages is used to generate user scores, then those scores are reweighted to better represent county demographics and ultimately aggregated to communities in time.

## Results

The depression and anxiety scores from our language based mental health assessments in 2020 adjusting for the past year can be found in Figure 3. The results as shown cover all weeks in 2020, and depict the included counties alongside the national average result. Assessments have been generated for all counties that demonstrate sufficient posting history to be considered reliable per the thresholds determined in this work, for this visualization a user threshold of 50 is used.

**CTLB Data Descriptives**

|  | Count |
|---|---:|
| Word Instances | 15,731,763,265 |
| Posts | 1,229,668,531 |
| Unique Words | 40,033,259 |
| Users | 2,045,124 |
| Counties | 1490 |
|  | **Mean (S.D.)** |
| Posts per User/Year | 161.78 (246.24) |
| Posts per User/Week | 6.97 (11.45) |
| Users per County | 1391.40 (4,859.35) |

**Table 1.** Coverage included in the filtered County Tweet Lexical Bank dataset from 2019 to 2020. We excluded English posts that were reposts and those that contained a hyperlink. Standard deviations are included next to average measurements.

**Reliability of Spatio-Temporal Resolutions.** Figure 2 shows the relationship between different resolutions of time and space on the split-half reliability of our measurements. Underlying all measurements we use depression scores within the given spatio-temporal cohorts. The threshold (Cohen's d = 0.1) was crossed for all township-level measurements, all but one county-level measurement, and all of the MSA-level measurements. Looking across time for counties we determine that the week level is the smallest time resolutions with our smallest accepted space resolution to have a reliability (1 - Cohen's d) that is ≥ 0.9. Using this county-week finding we observed that once there were at least 50 users (user threshold [UT]=50), reliability exceeded 0.8. In this context, the UT can be understood as the minimum number of unique users needed by a county to be included in our analysis. At a UT of 200 it is possible to obtain a reliability measurement of 0.9 indicating no effect. This analysis lead us to create standard county-week threshold guidelines at UT of 50 and 200. The use of the 50 UT reflects the higher number of counties that are directly usable versus the 200 UT, 809 counties for 50 UT and 411 counties for 200 UT.

**Convergent Validity.** Figure 4 depicts the outcomes of our multi-level fixed effects model between Gallup self-reported sadness and worry against our language based assessments of depression and anxiety. At all levels evaluated for fixed effects we find our t-test p value to be significant to 0.01. At the national level, we find that the survey and language are correlated (Pearson's r = 0.39) with depression and sadness, and anxiety and worry (r = 0.68). Fixed-effects coefficients between survey and language findings indicate higher agreement in analyses using larger spatial and temporal units, with the highest coefficients coming from a national-week analysis. At finer resolutions we nevertheless still identify positive values leading us to conclude that county-week level measurements may reflect greater local sensitivity that might not as consistently correspond to the greater national trends.

**External Criteria.** In Figure 5 we graphically represent the validity of our measures against other established county measures. The source of external county level data is the County Health Rankings (24) which track political, economic, societal, and health outcomes on a county-year scale. We observe strong agreement between the correlations of our LBMHA scores and the Gallup self-reported results with these PESH variables.

In Figure 6 we examine the difference between event weeks and non-event weeks. We find an increase on average of the mean absolute difference of both depression (23%) and anxiety (16%) during weeks in which major US events occur. Likewise we see a "resetting" effect wherein non-event weeks on average decrease the general level of both anxiety (6%) and depression (8%), however nationally across 2020 the absolute unadjusted level of both measures is increasing. These results over a comparison of event and non-event weeks for several counties suggest that changes in community mental health can be attributed to specific events.

**American Communities Comparison.** Figure 7 shows how anxiety differs across American community types. We select the five communities for which we the greatest representation in our final dataset of county-week LBMHAs. We observe that the Exurbs, defined as communities that "lie on the fringe of major metro areas in the spaces between suburban and rural America", score as the most anxious and most depressed of observed U.S. communities. Although overall difference between community types are modest, we anticipate that examinations of factorized measures of anxiety and depression may show larger discrepancies.

**Reliability per Spatiotemporal Unit**

|             | MSA (1) | County (62) | Township (370) |
|-------------|---------|-------------|----------------|
| Year (2)    | 0.993   | 0.933       | 0.802          |
| Quarter (3) | 0.996   | 0.948       | 0.816          |
| Month (8)   | 0.987   | 0.938       | 0.753          |
| Week (36)   | 0.986   | **0.921**   | 0.765          |
| Day (252)   | 0.977   | 0.888       | 0.684          |

**(a)** Reliability by spatial and temporal units for LBMHAs.

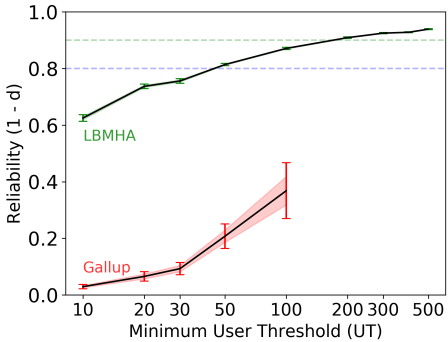

**(b)** Reliability vs. Minimum User Threshold for All County-Weeks

**Counts as Function of Minimum User Thresholds**

|                   | n > 200       | n > 50        | Full          |
|-------------------|---------------|---------------|---------------|
| County-Weeks      | 36,260        | 72,928        | 150,670       |
| Distinct Counties | 370           | 720           | 1,490         |
| Distinct States   | 51            | 51            | 51            |

**Means (S.D.) for County-Weeks**

|                   | n > 200       | n > 50        | Full          |
|-------------------|---------------|---------------|---------------|
| Users/County-Week | 1,585 (3,042) | 815 (2,297)   | 399 (1,650)   |
| Depression Score  | 2.41 (0.076)  | 2.42 (0.098)  | 2.42 (0.34)   |
| Anxiety Score     | 2.74 (0.073)  | 2.74 (0.097)  | 2.76 (0.35)   |

**(c)** County-Week Data Descriptives

**Fig. 2.** Spatiotemporal reliability of language based mental health assessments across different granularities of space and time. The heatmap in Table 2a shows the $1 - $ Cohen's d reliability of select New York metropolitan data, at each space and time unit $\geq 20$ unique users were required. From this heatmap we target the first entry greater than 0.9, which is county-week. The plot in Figure 2b shows how the reliability of a county-week measurement scales with the number of unique users that are required to include that county-week. Horizontal lines are drawn at 0.8 and 0.9 reliability, which were used to select a 50 and a 200 county user threshold. Standard error of the reliability is shown with red shading, and the 95% confidence interval is shown with error bars. The county-year Intraclass Correlations, test-length corrected (ICC2; (23)) at a UT of 50 are $ICC2 = 0.33$ for Gallup Sadness and $ICC2 = 0.97$ for LBMHA depression, while at a UT of 200 are $ICC2 = 0.87$ for Gallup and $ICC2 = 0.99$ for LBMHA. Table 2c shows data descriptives for the county-week dataset after applying a UT of 50 and 200 as per the reliability findings and all other thresholds.

## Discussion

Anxiety and depression are costly, underdiagnosed and undertreated, and while common overall, their prevalence varies across time and location. The present study used 15.7 billion words from by 2.05 million people living across the U.S. to evaluate a modern approach for measuring public mental health, using behavioral patterns (language use). We found this approach achieves much greater regional and temporal resolution (e.g., within large U.S. counties each week) while also achieving high convergent validity for the limited amount of high resolution survey-based assessments available.

Using recently develop computational methods to mitigate epidemiological selection bias (20), we establish minimal sampling thresholds, that could reliably and objectively describe the distribution and changes in public mental health (e.g., degrees of depression and generalized anxiety), both before and during the COVID-19 pandemic. We found that our pipeline reported similar temporal patterns, both nationally and at the county, to existing U.S. weekly data from Gallup but that we could report reliable results for a far larger number of counties and weeks.

Mental illnesses are suspected to be stigmatized because there are no highly sensitive biomarkers for the disease and, therefore, symptom presence or severity cannot be readily corroborated. To improve this process, this work join a recent line of work focused focused on identifying objective measures for mental illness. To date, some proposing that functional (25) or structural neuroimaging (26) might provide such a biomarker, while others have proposed that cellular changes might be more effective (27). Instead of relying on putative biomarkers to identify behavioral disorders, this study instead objectively determines levels of depression and anxiety by naturally observing individuals' unedited communications. In using one of the only databases that regularly records people's written reactions to everyday events by individuals residing across the US, this study objectively monitored how people interpreted and processed local and global events. Future work will expand on this process to both refine the tool and improve the generalizability as we move this work into public mental health monitoring programs.

In this study, we objectively measured mental health to answer the question: what is happening for mental health here and now? This shared geographic and temporal resolution is a leap forward in our ability to understand the role of social, economic, or natural events and mental health.

For example, this study shows that improved resolution of mental health outcomes reflect the presence of a major national events. For example, following the murder of George Floyd, language-estimated depression prevalence showed a clear increase, mirroring similar trends observed in Gallup survey data (28).

COVID-19 first arrived in the U.S. during the data collection period (2019 to 2020). Consistent with prior research, we found that COVID-19 caused a rapid increase in depressive symptoms and generalized anxiety across the U.S. that did not dissipate before 2021. The distribution of poorer mental health was widespread and included large increases in regions with relatively low pre-pandemic levels of depression and anxiety. For example, the average level of anxiety increased from the lowest to the highest levels in Kansas in the months after the pandemic. These mental health shocks also began late in 2019, when COVID-19 was first being identified globally, and spiked in early March 2020 when much of the North Eastern U.S. was locked down and many people in states that remained open chose to self-isolate. While these effects show the value of the approach for understanding how public mental health changes in a pandemic, these data also show that anxiety and depressive symptoms had not yet returned to pre-pandemic norms by the end of the observational window.

We observed mental health using posts from geo-located Twitter users, as this allowed us to examine rapid changes in mental health at scale. LBMHAs have been reliably used outside of social media. For example, studies of psychological stress have noted that LBMHA can aid in identifying individ-uals with at risk of poorer postpartum mental health when relying on mothers' diaries (29) and for identifying poorer long-term prognosis in post-traumatic stress disorder when relying on oral histories (19).

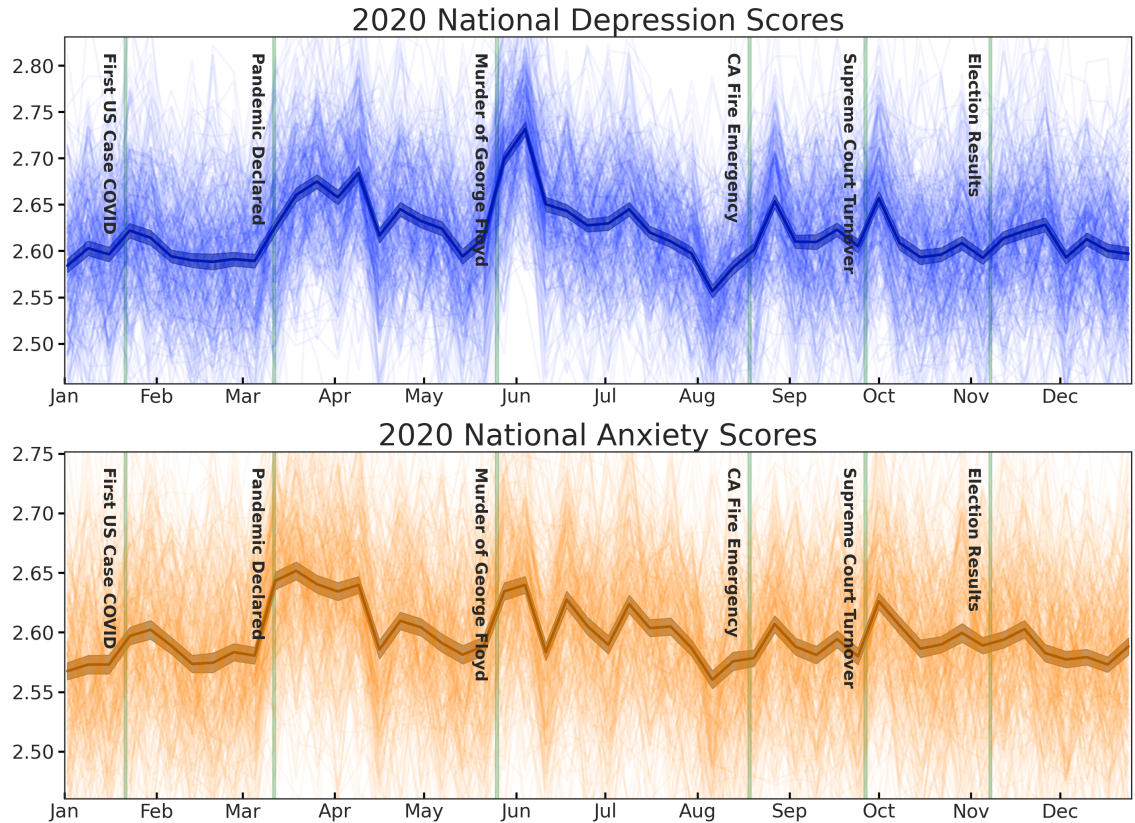

**Fig. 3.** Depression (blue) and anxiety (orange) measured at the nation-week level for all of 2020, controlling for 2019 measurements. All scores shown are based on aggregated user scores that are scaled from 0 to 5, 5 representing the highest level of depression/anxiety. Labeled green vertical markers are placed on the start of major events. In dark blue/orange, we have plotted nation-week averages alongside x 95% confidence intervals, and in thinner lines we show similar trends for individual counties. This figure requires counties to contain at least unique 200 (UT=200) users in a given week to be included, this gives 370 counties covering all weeks in 2020.

**Limitations.** Results from this study should be interpreted in light of a number of limitations. First, many U.S. counties with small populations or a small numbers of social media users were not sufficiently large to provide reliable estimates. These areas, though numerous, have relatively small populations and account for a small percentage of the U.S. population. Nevertheless, they are regions that are often under-represented in research studies and more should be done to try to include information from these regions.

Additionally, social media platforms and their populations change based on their policies and ownership. Twitter recently changed ownership resulting in a noticeable change in content and new data sharing practices. While other sources of public language exists, the evaluations of this paper are limited to prior years of Twitter and any application to other platforms or after the recent ownership change require further validation. This work mainly centers around the analysis of 2020 LBMHAs and utilizes 2019 as a control.

Still, the strength of this epidemiological study is that it applied scalable methods meant to improve generalizability on a sample that included >900 million observations on 2 million individuals (0.6% of the U.S. population) across more than 1,400 U.S. counties. These results are to our knowledge the first to validate temporal results only previously available from U.S. polling sites interested in tracking mental health.

**Implications for Population Health.** To date, most efforts to profile the mental health of people in the U.S. and globally are relying on subjective information provided to survey prompts. These may biased by the tendency for people to under-report less desirable or stigmatized traits, such as the presence of mental illness. Up to date access to objective measures of changing mental health could improve in the ability to allocate scarce mental health treatment resources in a time of great need, and will facilitate new analyses that can help us to better understand the risk factors and consequences of depression and anxiety in the population.

## Materials and Methods

**2019–2020 County Tweet Lexical Bank.** As our main source of social media data we introduce an updated version of the original *County Tweet Lexical Bank* (30) which we refer to as *CTLB-19-20*. This new version contains a cohort of county mapped Twitter accounts and their posts spanning 2019 from to 2020. These county-user pairs were derived form posts with either explicit longitude/latitude pairs or the first instance of a self-reported user location in the account public profile. Previous work mapping location strings to counties was found to be 93% accurate compared to human assessments (31). The

**Convergent Validity: Language and Survey**

| Space ($N$) | Time ($N$) | Depression $\beta$ | Anxiety $\beta$ |
|---|---|---|---|
| National (1) | Week (22) | 0.583‡ | 1.582‡ |
| Regions (4) | Weeks (22) | 0.613‡ | 1.533‡ |
| Counties (132) | Quarters (3) | 0.346‡ | 1.178‡ |
| Counties (132) | Weeks (22) | 0.254‡ | 0.371‡ |

**(a)** Association by Space and Time Resolutions

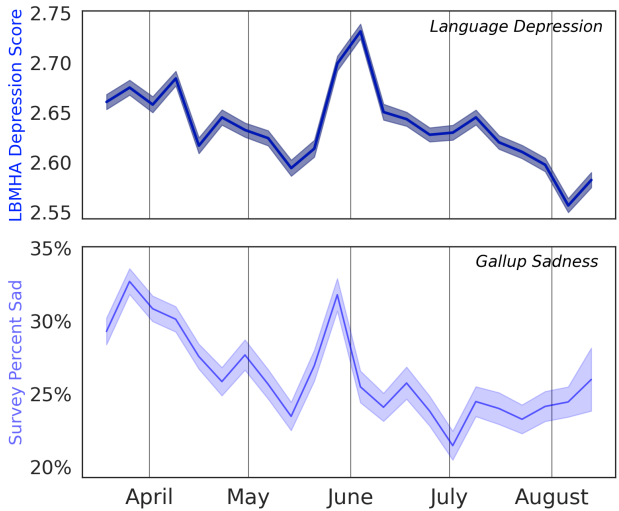

**(b)** Language and Surveys across 2020, Nationally

**Fig. 4.** Convergent validity (hierarchical linear modeling coefficient) between language-based assessments and survey-based measures at different spatial and temporal resolutions. Depression compares our language-based scores to Gallup's surveyed sadness scores. Anxiety compares our language-based anxiety scores to surveyed Gallup's worry scores. Table 4a shows fixed-effects coefficients between language based mental health assessments and measurements collected by the Gallup COVID-19 Panel Questionnaire. Figure 4b shows the national plots of depression as measured by LBMHAs and sadness as measured by Gallup, with fixed effects coefficients are provided in the upper right. Both Questionnaire and LBMHA measures are held to reliability constraints as described in our section on reliability. Between the national-week two plots shown there is a $\beta = 0.582$.
Results significant at: ‡$p < .001$, †$p < .01$, *$p < .05$

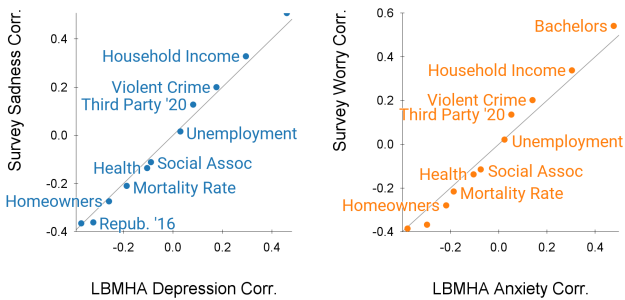

**Fig. 5.** Cross-sectional associations between language based mental health assessments (LBMHAs) of Anxiety/Depression and survey based assessments of Worry/Sadness against external criteria from Political, Economic, Social, and Health (PESH) variables across $N = 132$ counties. Scatterplots of correlations between external criteria and our scoring method on one axis and the surveyed results on the other axis. All counties included meet our reliability requirements. Shown in a diagonal gray line is $y = x$ which would indicate perfect agreement. Association is measured using Pearson correlation and all findings are significant to $p < 0.05$.

unprocessed CTLB-19-20 contained 2.7 billion total posts from a cohort of 2.6 million accounts over 2019 and 2020 (see Table

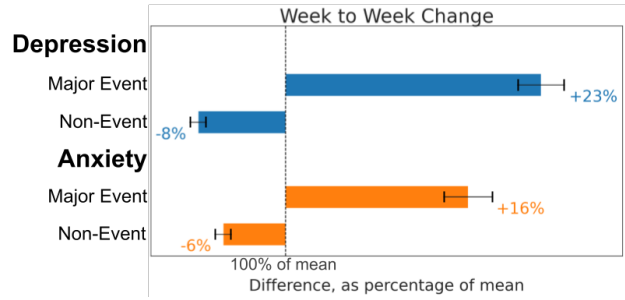

**Fig. 6.** Longitudinal analysis of the impact of weeks containing major US events against weeks without similar events. Shown are the z-scored percent differences from the prior week in LBMHAs between weeks that do contain major US events and those weeks that do not. Confidence interval bars are generated from Monte Carlo bootstrapping on 10,000 samples from the pool of either event weeks or non-event weeks and re-calculating mean z-scored percent differences between the drawn samples.

1). For each post in this dataset we retain the date it was posted, a unique user identifier, the original text body, and the US county that the poster is from.

Following Giorgi and colleagues (30), preprocessing steps filtered out posts to increase the accuracy of social media based population assessments (32). Posts are only included if they are marked likely to be English according to the langid package (33), and then they are further filtered to remove reposts, posts containing URLs (i.e. posts likely of non-original content), and finally any duplicate messages from individual users. The final processed dataset contains nearly 1 billion posts across of 2 million unique accounts for all 104 weeks in 2019 and 2020. At this point 1,490 counties (whose total population equals ∼92.5% of the US population) are captured. Further statistics about the filtered CTLB are described in more detail in Figure 1.

To maintain a minimum level of reliability for our depression and anxiety measurements users must post at least 3 times in a given week to be included in that week, and from our reliability testing we determined that counties must contain at least 200 unique users per week to be considered for any given week. The 3 user posting threshold (3-UPT) was determined to balance diversity of users while minimizing noise from infrequent users. The 3-UPT approach resulted in a 37% decreased in unique user-week pairs retained, as opposed to a 23.4% decrease for 2-UPT and a 53% loss for 5-UPT. The 200 user threshold (UT) was determined by a reliability analysis whose results are shown in Table 2. Counties that fail to report a score for 10 weeks consecutively are dropped from the dataset to remove the influence they pose to findings for a single week.

After applying our 3-UPT, UT, and max gap filtering many, mostly rural, counties are necessarily excluded from our analysis. Since the target of this work is to better meet mental health reporting needs we implement a super-county binning strategy to reincorporate those "unreliable" county findings. All county-week findings that fail to meet the UT filter are weighted-mean aggregated by state into a super county-week result. Weights for the mean aggregation are assigned based on the reporting population of users of the included counties. Super counties must then pass the same UT set for regular counties to be included. In the case of UT=200 this results in a gain of 4,714 super county-week results over the original 30,899 county-week results. Figure 7

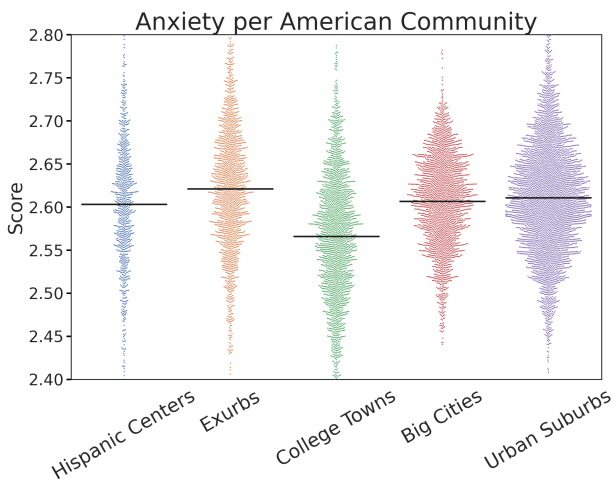

(a) American Community anxiety scores

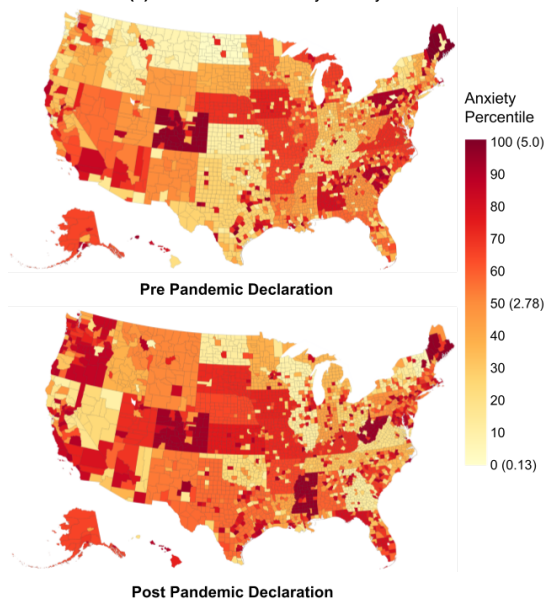

(b) Map of anxiety before and after pandemic declaration

**Fig. 7.** Scores within communities in 2020 and county mapped anxiety before and after COVID-19 is declared a pandemic. In 7a communities, as defined by the American Communities Project, are shown ranked from highest to lowest mean measurement captured. A mean line is overlaid on swarm plots of the county-week measurements for each community type. The five communities shown were chosen because they had the largest numbers of weekly LBMHAs in our data. In 7b 20-percentile county-level measurements of anxiety are shown, where red shows where anxiety is most prevalent and yellow where anxiety is least prevalent. Pre-declaration is defined as January-February 2020 and post-declaration is defined as April-May 2020. Super-county binning is performed to report results for counties that are not individually reliable.

visually demonstrates how super-county binning reincorporates findings from unreliable counties.

The final post-processing step in our county-week pipeline is to run linear interpolation on a per county basis between missing weeks. For reference, at UT=200 this translates to an increase from 35,613 to 36,260 county-weeks. When running our analyses in this work we opt to adjust 2020 county-week findings by removing periodicity effects by subtracting means for 2019. This adjustment highlights 2020-specific movement from week to week.

**Language Based Assessments.** To extract language based assessments of well-being from posts, we used existing lexical models of depression and anxiety (34, 35) that we adapted to 2019-2020 Twitter vocabularies using target-side domain adaptation (36) which removes lexical signals that have different usage patterns (see target domain adaptation). The process for applying the model consists of extracting words from posts using the social media-aware tokenizer from *dlatk* (37). Following (38), the relative frequency of the words per user and unit of time are then Anscombe transformed to stablize the variance of power law distribution. The approach then applies a linear model that is pretrained to produce anxiety and prediction scores from the word frequencies (35, 39). This produces a degree of depression (DEP_SCORE) and degree of anxiety (ANX_SCORE) for each user-week pair in the processed dataset.

The calculation of a langauge based mental health scoring, for example the depression score for a user-week, is defined as:

$$LBMHA_{DEP}(x) = L(x) \times \text{demographics}(x)$$

$$L(x) = \sum_{w \in lex} [(A_{ns}(freq_w(x))) \times lex_{wt}(w)] + lex_i(DEP)$$

where $LBMHA_{measure}(x)$ is the Language Based Mental Health Assessment of an entity in time. $x$, is the sum of the summation of the lexicon weights $lex_{wt}()$ of all words $w$ in the lexicon $lex$ times that word's Anscombe transformed frequency, $A_{ns}(freq_w())$, and the overall lexicon intercept $lex_i()$ for that particular assessment. This outcome is multiplied by demographics(), which maps to a per user-week post-stratified weight correcting for the socio-economics of the community (see post-stratification).

It is noted that Twitter is a biased sample of the American populace, we find that their users are younger, more educated, and more male than the average American (40). In order to correct for these discrepancies from the true socioeconomic diversity of US counties we apply a weighting scheme to emphasize the language of voices that are under-represented in tweets. Post-stratified weights are an effective method for accomplishing this as described by (20). Our generated user-week level weights are applied onto the language based assessments of depression and anxiety calculated thus far.

The final reported user-week scores for depression and anxiety are then clipped to be between 0 and 5 for ease of interpretation. From these final scores, weighted aggregates can be generated at higher space and time resolutions.

**Target Domain Adaptation.** The mental health lexicon used in this work was originally trained for use on Facebook posts in the late 2000s so the following target-side domain adaptation steps were taken to adapt the lexicon to Twitter language in 2019-2020. In comparing the language use of Facebook versus Twitter we first trimmed the original lexicon's vocabulary which contained 7,680 unique words, to a set of 5,765 words for the target set where the *word usage* and *mean word frequency* between the two domains fell within certain ranges of each other.

The usage and frequency filters were used to address the phenomena of words and phrases that are used with different frequencies between two domains of text being more likely to have significant differences in their semantics between those two domains (36, 41). As the correlation between the frequency of a phrase and outcomes for a given lexicon may differ for semantically different usages of a phrase, filtering words with different usages and frequencies limits our set of tokens to those that are more likely to carry similar semantics (and thus, similar correlations). We modify (36)'s frequency filter for the source to target adjustment to instead normalize by standard deviation across the source Facebook users, and introduce a usage filter (what percent of users in each domain used a specific token even once).

Specifically, for each of our two domains (the target Twitter domain and the source Facebook domain), we computed each user's frequency for each word, and stored the results in frequency matrices $C^S$ of dimension $n \times m$ and $C^T$ of dimension $k \times m$, where $n$ is the number of users in our source domain, $k$ is the number of users in our target domain, and $m$ is the cardinality of the set of words that appear either in the Twitter or Facebook domain. For each word, we then computed the average relative frequency across all users (word frequencies $f^S$ for Facebook and $f^T$ for Twitter), and the percent of users who used the word at least once (word usage percentages $u^S$ and $u^T$).

First, only words with word usage percentages within a multiplicative factor 10 across domains were kept ($-1 < \log_{10}(u^T/u^S) < 1$), leaving 6,214 words. Then, for each word we take a Cohen's $d$ filter of $f^S$ versus $f^T$ in the range $[-0.2, 0.2]$ on the word frequency using the larger source domain's standard deviation. A mathematical definition of this process is given in the supplement materials.

Finally we dropped common US names found in the United States' Social Security list of Popular Baby Names by Decade (e.g. Emma, Noah, Olivia, Liam)(42). The resulting Twitter adapted lexicon vocabulary after these three filters is 5,469 words long.

Using the Differential Language Analysis ToolKit's (DLATK) (37) regression-to-lexicon feature a new lexicon was trained using ridge regression, we note that the option to not standardize is selected since it better suits the lexicon creation task.

To adapt lexical patterns to the target domain, we remove words which display different usage patterns in the target domain. Specifically, words that appeared with significantly different distributions in terms of sparsity or mean frequency. We then retrain the lexical model of (34, 35) based on this filtered set of words to generate our domain-adapted well-being lexica (36).

The final lexicon contained 5,765 words and an intercept each with a weight for depression (DEP_SCORE) and anxiety (ANX_SCORE).

## Statistical Analysis

**Reliability of Spatio-Temporal Resolutions.** At this point, we can begin to aggregate to a larger spatial or temporal resolution as necessary for analysis. To determine an appropriate resolution, we examine the finest resolution we can achieve while retaining reliable depression and anxiety score measurements.

To evaluate the reliability of a given spatio-temporal resolution, for each space-time pair in the resolution, we gather the set of users who posted at least 3 messages in this time period. If there are at least 20 such users, we randomly split the set into two approximately equally sized subsets and compute the split-half reliability ($R = 1 - $ Cohen's d) using their depression scores. Finally, the reliability is averaged across all space-time pairs.

Figure 2 shows the reliability scores of different spatio-temporal resolutions from running the procedure with counties in the New York City metropolitan area.

It is possible to generate reliable measures ($R > 0.9$) at the county-week level. We also analyze the effect of the threshold for the number of users per county-week pair on reliability. Figure 2 shows the reliability scores from running the aforementioned procedure with the entire CTLB data and with different thresholds for the number of users.

When relying on regional data, we report data that exceed a final group frequency threshold placed at 50 or 200 to match repeated split-half reliability (RSR) where RSR $> 0.7, 0.8$, and $0.9$ for these thresholds respectively. $RSR$ is calculated as the mean Cohen's d of $N$ repeated split-half samples into equal length $a$ and $b$ halves from the data belonging to a given region in time.

$$RSR = \frac{1}{N} \sum_{i=1}^{N} 1 - \frac{\mu_a - \mu_b}{\sigma_{a \cup b}}$$

**Convergent Validity.** Figure 4 we look to the Gallup COVID Panel (43) to compare the validity of our measure and determine if these assessments are tracking the same underlying construct. Note that we do not treat the Gallup poll as a gold standard to exactly align with since the poll is a survey based measure of self-reported sadness and worry, while our language based assessments are scores of depression and anxiety. The purpose of this particular study is to show common alignment between a traditional survey method and an observational social media method. The Gallup data is based on individual responses to a survey which are then tagged with a week and a county of the respondent. This dataset covers 2617 counties with an average of ∼4,601 measurements per week across all counties. To this end we use fixed effect multi-level modeling to remove the effects of endogeneity bias stemming from inherent between-county differences. While LBMHA scores are already held to a baseline 1-Cohen's d reliability of 0.9, Gallup results are held to a standard of 0.7. If this adjustment is not made there are no counties collected by Gallup for which county-week results are reliable for the full 22 weeks the survey covered.

**External Criteria.** To compare our assessments cross-sectionally against other external measurements we look to the County Health Rankings (CHR) (24). From CHR 2020 we look to political, economic, social, and health based outcomes at the county level. For political variables we evaluate the proportion of county voters who voted Republican in 2016 and 2020 and Third party in 2020. For economic variables, the logged median household income, the unemployment rate, and the proportion of homeowners. For social variables, the per capita number of social associations, the violent crime rate, and the percent of youth unaffiliated with school or a similar organization. For health variables, the surveyed percent of people reporting fair or poor health, and the age-adjusted mortality rate. Figure 5 extends the cross-sectional test of validity to conduct a longitudinal study of major events on measurements across counties. For this work we examine the weekly changes in county measurements of anxiety and depression during weeks where major US events occurred and weeks where they did not occur. Combining 14 events identified by The Uproar (44) with 18 events from Business Insider (45) we arrived at 14 weeks of 2020 as "major US event weeks" (13 events were in common between the news sources and a single week could contain more than 1 event). We then filtered these to those that happened within the United States (including those applying global, such as pandemic onset) arriving at 14 total event weeks to compare with 38 non-event weeks. An event week is defined as an ISO week which contains the date any of the labelled major events occurred on. A 1 day buffer is added to the date of the event before mapping to a week so that scoring changes caused by the event can be captured. For each sample of event and non-event weeks, we collect the percent change in national-week depression and anxiety scores from the previous week. Using these two samples we compute Cohen's d between the event week and non-event week findings. To establish a confidence interval we use Monte Carlo bootstrapping over 10,000 iterations of event and non-event weeks.

**Data Sharing and Availability.** To support open science, we provide an open-source toolkit to run the LBMHA pipeline as well as data describing the results per county week. Please see LINK for a repository of code and data associated with this article.

**ACKNOWLEDGMENTS.** Support for this work was provided by a grant by Joint The National Institutes of Health and National Science Foundation Program, Smart and Connected Health, (Grant NIH/NIMH R01 MH125702, PIs Eichstaedt, Schwartz.) as well as a CDC/NIOSH Grant U01 OH012476.

1. S Abuse, MHS Administration, Key substance use and mental health indicators in the united states: results from the 2019 national survey on drug use and health. *HHS Publ. No* **52**, 17–5044 (2020).
2. AJ Baxter, T Vos, KM Scott, AJ Ferrari, HA Whiteford, The global burden of anxiety disorders in 2010. *Psychol. Medicine* **44**, 2363–2374 (2014).
3. HA Whiteford, et al., Global burden of disease attributable to mental and substance use disorders: findings from the global burden of disease study 2010. *The lancet* **382**, 1575–1586 (2013).
4. EA Knapp, U Bilal, LT Dean, M Lazo, DD Celentano, Economic insecurity and deaths of despair in us counties. *Am. journal epidemiology* **188**, 2131–2139 (2019).
5. A Case, A Deaton, *Deaths of Despair and the Future of Capitalism*. (Princeton University Press), (2021).
6. Y Milaneschi, WK Simmons, EF van Rossum, BW Penninx, Depression and obesity: evidence of shared biological mechanisms. *Mol. psychiatry* **24**, 18–33 (2019).
7. MA Davis, LA Lin, H Liu, BD Sites, Prescription opioid use among adults with mental health disorders in the united states. *The J. Am. Board Fam. Medicine* **30**, 407–417 (2017).
8. P Nsubuga, et al., Public health surveillance: a tool for targeting and monitoring interventions. *Dis. Control. Priorities Dev. Countries. 2nd edition* (2006).
9. G Rose, Sick individuals and sick populations. *Int. journal epidemiology* **30**, 427–432 (2001).
10. Gallup, Health rating remains below pre-pandemic level [internet] (2021).
11. J Hsia, et al., Comparisons of estimates from the behavioral risk factor surveillance system and other national health surveys, 2011- 2016. *Am. journal preventive medicine* **58**, e181–e190 (2020).
12. NIoMH NIMH, *Prevalence of Generalized Anxiety Disorder Among Adults*. (National Institutes of Health, Bethesda, MD), (2021).
13. JT Chen, N Krieger, Revealing the unequal burden of covid-19 by income, race/ethnicity, and household crowding: Us county versus zip code analyses. *J. Public Heal. Manag. Pract.* **27**, S43–S56 (2021).
14. N Krieger, et al., Geocoding and monitoring of us socioeconomic inequalities in mortality and cancer incidence: does the choice of area-based measure and geographic level matter? the public health disparities geocoding project. *Am. journal epidemiology* **156**, 471–482 (2002).
15. AL Kratz, SL Murphy, TJ Braley, Ecological momentary assessment of pain, fatigue, depressive, and cognitive symptoms reveals significant daily variability in multiple sclerosis. *Arch. physical medicine rehabilitation* **98**, 2142–2150 (2017).
16. MA Russell, JM Gajos, Annual research review: Ecological momentary assessment studies in child psychology and psychiatry. *J. Child Psychol. Psychiatry* **61**, 376–394 (2020).
17. MJ Paul, M Dredze, Social monitoring for public health. *Synth. Lect. on Inf. Concepts, Retrieval, Serv.* **9**, 1–183 (2017).
18. K Jaidka, et al., Estimating geographic subjective well-being from twitter: A comparison of dictionary and data-driven language methods. *Proc. Natl. Acad. Sci.* **117**, 10165–10171 (2020).
19. Y Son, et al., World trade center responders in their own words: predicting ptsd symptom trajectories with ai-based language analyses of interviews. *Psychol. medicine* **2021 Jun 22**, 1–9 (2021).
20. S Giorgi, et al., Correcting sociodemographic selection biases for population prediction from social media in *Proceedings of the International AAAI Conference on Web and Social Media*. Vol. 16, pp. 228–240 (2022).
21. AP Christie, et al., Quantifying and addressing the prevalence and bias of study designs in the environmental and social sciences. *Nat. communications* **11**, 1–11 (2020).
22. J Mellon, C Prosser, Twitter and facebook are not representative of the general population: Political attitudes and demographics of british social media users. *Res. & Polit.* **4**, 2053168017720008 (2017).
23. PD Bliese, Within-group agreement, non-independence, and reliability: Implications for data aggregation and analysis. *Multilevel theory, research, methods organizations* (2000).
24. U of Wisconsin Population Health Institute, County health rankings and roadmaps 2022. (2020).
25. JR Sato, et al., Machine learning algorithm accurately detects fmri signature of vulnerability to major depression. *Psychiatry Res. Neuroimaging* **233**, 289–291 (2015).
26. M Kritikos, et al., Cortical complexity in world trade center responders with chronic posttraumatic stress disorder. *Transl. Psychiatry* **11**, 1–10 (2021).
27. PF Kuan, et al., Metabolomics analysis of post-traumatic stress disorder symptoms in world trade center responders. *Transl. psychiatry* **12**, 1–7 (2022).
28. JC Eichstaedt, et al., The emotional and mental health impact of the murder of george floyd on the us population. *Proc. Natl. Acad. Sci.* **118**, e2109139118 (2021).
29. A Bartal, KM Jagodnik, SJ Chan, MS Babu, S Dekel, Identifying women with postdelivery posttraumatic stress disorder using natural language processing of personal childbirth narratives. *Am. J. Obstet. & Gynecol. MFM* **5**, 100834 (2023).
30. S Giorgi, et al., The remarkable benefit of user-level aggregation for lexical-based population-level predictions in *Proceedings of the 2018 Conference on Empirical Methods in Natural Language Processing*. (Association for Computational Linguistics), pp. 1167–1172 (2018).
31. H Schwartz, et al., Characterizing geographic variation in well-being using tweets in *Proceedings of the International AAAI Conference on Web and Social Media*. Vol. 7;1, pp. 583–591 (2013).
32. K Jaidka, et al., Estimating geographic subjective well-being from twitter: A comparison of dictionary and data-driven language methods. *Proc. Natl. Acad. Sci.* **117**, 10165–10171 (2020).
33. M Lui, T Baldwin, langid. py: An off-the-shelf language identification tool in *Proceedings of the ACL 2012 system demonstrations*. pp. 25–30 (2012).
34. HA Schwartz, et al., Towards assessing changes in degree of depression through facebook in *Proceedings of the workshop on computational linguistics and clinical psychology: from linguistic signal to clinical reality*. pp. 118–125 (2014).
35. Y Son, et al., World trade center responders in their own words: predicting ptsd symptom trajectories with ai-based language analyses of interviews. *Psychol. Medicine* p. 1–9 (2021).
36. D Rieman, K Jaidka, HA Schwartz, L Ungar, Domain adaptation from user-level facebook models to county-level twitter predictions in *Proceedings of the Eighth International Joint Conference on Natural Language Processing (Volume 1: Long Papers)*. pp. 764–773 (2017).
37. HA Schwartz, et al., Dlatk: Differential language analysis toolkit in *Proceedings of the 2017 conference on empirical methods in natural language processing: System demonstrations*. pp. 55–60 (2017).
38. HA Schwartz, et al., Personality, gender, and age in the language of social media: The open-vocabulary approach. *PloS one* **8**, e73791 (2013).
39. HA Schwartz, et al., Predicting individual well-being through the language of social media in *Biocomputing 2016: Proceedings of the Pacific Symposium*. (World Scientific), pp. 516–527 (2016).
40. G Blank, C Lutz, Representativeness of social media in great britain: investigating facebook, linkedin, twitter, pinterest, google+, and instagram. *Am. Behav. Sci.* **61**, 741–756 (2017).
41. P Resnik, Using information content to evaluate semantic similarity in a taxonomy in *Proceedings of the 14th International Joint Conference on Artificial Intelligence - Volume 1*, IJCAI'95. (Morgan Kaufmann Publishers Inc., San Francisco, CA, USA), p. 448–453 (1995).
42. S Security, Popular baby names by decade (year?).
43. Gallup, Covid-19 panel microdata (2021).
44. C Majerac, The 14 most important events of 2020. *The Uproar: https://nashuproar.org/39777/features/the-14-most-important-events-of-2020* (2020).
45. Y Dzhanova, The events that shook and shaped america in 2020. *Bus. Insid. https://www.businessinsider.com/the-stories-of-2020-that-shaped-and-shook-americans-2020-12* (2020).

## Supplemental Materials

**Target Domain Adaptation.** In retraining our ridge regression to generate a well-being lexicon, a k-fold analysis determined that the best alpha to use on the ridge regression was 0.001, as well as preprocessing the data down to k=500 components using principal component analysis.

For our domain adaptation with word usage and frequency-based, we used DLATK to get each user's group norm for a given token (the relative frequency of a token across all the user's text, i.e., the sum of word $j$'s count divided by the total number of word's in a user's corpus) and store those in sparse matrices $C_{i,j}^S$ (of size $n \times m$) for Facebook and $C_{i,j}^T$ (of size $k \times m$) for Twitter, where the rows represent the number of user's in each domain and the column's represent unique identifiers for any word that appears at least once in one of the two datasets. Entry $C_{i,j}^S$ is equal to the number of times user $i$ in the source domain, used token $j$ across all of their posts (while $C_{i,j}^T$ is analogous for user $i$ in the target domain).

The Target mean word frequency $f^T$ and Source mean frequency $f^S$ are defined as the averages for each word with index $j$:

$$f_j^S = \frac{1}{n} \sum_{i=1}^{n} C_{i,j}^S$$

$$f_j^T = \frac{1}{k} \sum_{i=1}^{k} C_{i,j}^T$$

while the Source standard deviation is analogously defined as the standard deviation across the set of source users.

$$\sigma_j^S = \frac{1}{N} \sqrt{\sum_{i=1}^{N} \left( S_{i,j} - f_j^S \right)^2}$$

Meanwhile, the target and source word usage for token $j$ are defined as the fraction of users that use token $j$ at least once. These are computed by letting $B^T$ and $B^S$ be the binary usage matrices corresponding to $T$ and $S$:

$$B_{i,j}^S = \begin{cases} 1, & C_{i,j}^S > 0 \\ 0, & \text{otherwise} \end{cases}$$

$$B_{i,j}^T = \begin{cases} 1, & C_{i,j}^T > 0 \\ 0, & \text{otherwise} \end{cases}$$

The word usage $u_j^S$ and $u_j^T$ for token $j$ are thus:

$$u_j^S = \frac{1}{n} \sum_{i=1}^{n} B_{i,j}^S$$

$$u_j^T = \frac{1}{k} \sum_{i=1}^{k} B_{i,j}^T$$

The filters chosen were a domain usage filter $log_{10}$ ratio between [-1.0,1.0] where the log ratio $L$ is computed as $log_{10}$(Target word usage / Source word usage), this left 6,214 remaining words:

$$L_j = \log_{10} \left( \frac{u_j^T}{u_j^S} \right)$$

Then a frequency filter was applied from [-0.2,0.2], calculated as (Target mean word Frequency - Source mean word frequency) / Source standard deviation of word frequency):

$$d_j = \frac{f_j^T - f_j^S}{\sigma_j^S}$$

Siddharth Mangalik, Johannes C. Eichstaedt, Salvatore Giorgi, Jihu Mun, Farhan Ahmed, Gilvir Gill, Adithya V. Ganesan, Shashanka Subrahmanya, Nikita Soni, Sean A. P. Clouston, H. Andrew Schwartz